\pdfoutput=1

\documentclass[11pt]{article}

\usepackage[final]{acl}

\usepackage{times}
\usepackage{latexsym}

\usepackage[T1]{fontenc}
\usepackage[utf8]{inputenc}

\usepackage{microtype}

\usepackage{inconsolata}

\usepackage{graphicx}

\usepackage{xspace}
\usepackage{amsfonts}
\usepackage{amsthm}
\usepackage{amsmath}
\usepackage{amssymb}
\usepackage{cleveref}
\usepackage{bbm}
\usepackage{caption}
\usepackage{subcaption}
\usepackage{xcolor}
\usepackage{scalerel}
\usepackage{booktabs}

\usepackage{amssymb}%
\usepackage{pifont}%
\newcommand{\cmark}{\ding{51}}%
\newcommand{\xmark}{\ding{55}}%

\usepackage[textsize=tiny]{todonotes}
\newtheorem{defin}{Definition}

\usepackage{cleveref}
\crefname{section}{\S}{\S\S}
\Crefname{section}{\S}{\S\S}
\crefname{table}{Tab.}{}
\crefname{figure}{Fig.}{Figs.}
\crefname{algorithm}{Alg.}{}
\crefname{equation}{Eq.}{}
\crefname{appendix}{App.}{}
\crefname{thm}{Theorem}{}
\crefname{prop}{Proposition}{}
\crefname{defin}{Definition}{}
\crefname{cor}{Corollary}{}
\crefname{observation}{Observation}{}
\crefname{assumption}{Assumption}{}
\crefformat{section}{\S#2#1#3}
\crefformat{footnote}{#2\footnotemark[#1]#3}

\newcommand{\colourbase}{purple}
\newcommand{\mymacro}[2]{\newcommand{#1}{{\color{\colourbase}#2}}}

\newcommand{\defn}[1]{\textbf{#1}}
\DeclareMathOperator*{\expect}{\mathbb{E}}

\mymacro{\ptrue}{p}
\mymacro{\btheta}{\boldsymbol{\theta}}
\mymacro{\bthetaprime}{\btheta'}
\mymacro{\ptheta}{p_{\scaleto{\btheta}{4pt}}}
\mymacro{\pthetaprime}{p_{\scaleto{\bthetaprime}{4.5pt}}}
\mymacro{\subword}{s}
\mymacro{\subwords}{\mathbf{s}}
\newcommand{\context}{\subwords_{\scaleto{<t}{4.5pt}}}
\newcommand{\contextplusword}{\subwords_{\scaleto{\leq t}{4.5pt}}}
\newcommand{\kl}{\mathrm{KL}}

\mymacro{\convergencefunc}{\mathtt{conv}}
\newcommand{\expconvergence}{\expect[\convergencefunc]}
\newcommand{\condexpconvergence}[1]{\expect_{#1}[\convergencefunc]}
\mymacro{\distfuncbase}{d}
\newcommand{\distfunc}[3]{\distfuncbase_{\scaleto{#1}{5pt}}(#2, #3)}
\mymacro{\vocab}{\mathcal{S}}
\mymacro{\nounset}{\vocab_{\mathtt{noun}}}
\mymacro{\verbset}{\vocab_{\mathtt{vrb}}}
\mymacro{\freqset}{\vocab_{\mathtt{freq}}}
\mymacro{\dataset}{\mathcal{D}}

\makeatletter
\DeclareRobustCommand*{\escapeus}[1]{%
  \begingroup\@activeus\scantokens{#1\endinput}\endgroup}
\begingroup\lccode`\~=`\_\relax
   \lowercase{\endgroup\def\@activeus{\catcode`\_=\active \let~\_}}
\makeatother
\newcommand{\myemph}[1]{\textsf{{\escapeus{#1}}}}

\usepackage{tabularray}
\usepackage{cellspace}
\newcommand\cincludegraphics[2][]{\raisebox{-0.2\height}{\includegraphics[#1]{#2}}}
\usepackage{setspace}

\title{Convergence and Divergence of Language Models \\
under Different Random Seeds}

\newcommand{\utexasid}{{\includegraphics[scale=0.015]{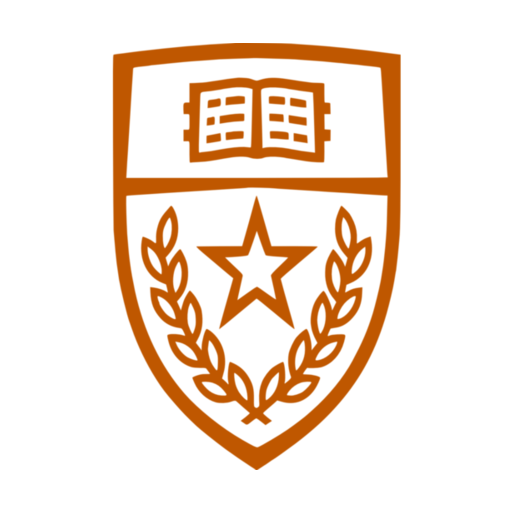}}}
\newcommand{\ethid}{{\includegraphics[scale=0.028]{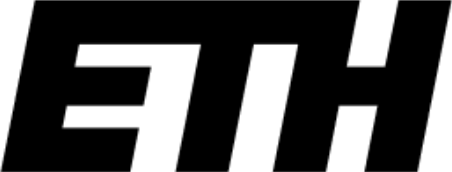}}}

\author{
    Finlay Fehlauer,$^\ethid$\,\,\,
    Kyle Mahowald,$^\utexasid$\,\,\,
    Tiago Pimentel\,$^\ethid$ \\ ~\\
    \textsuperscript{\ethid}ETH Z\"urich \quad
    \textsuperscript{\utexasid}University of Texas at Austin \\
    {\myemph{\href{mailto:ffehlauer@ethz.ch}{ffehlauer@ethz.ch}},\,\, \myemph{\href{mailto:mahowald@utexas.edu}{mahowald@utexas.edu}},\,\, \myemph{\href{mailto:tiago.pimentel@inf.ethz.ch}{tiago.pimentel@inf.ethz.ch}}} \\
    \begin{tblr}{colspec = {Q[c,m] Q[c,m]}, colsep=10pt, stretch=0}
        \cincludegraphics[width=1.1em, keepaspectratio]{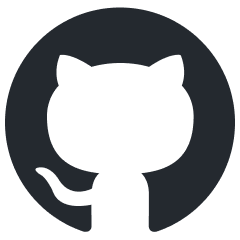} {\fontsize{11pt}{11.5pt}\selectfont\href{https://github.com/Tr1ple-F/convergence-and-divergence-of-llms}{\myemph{Tr1ple-F/convergence-and-divergence-of-llms}}}
    \end{tblr}
}

\begin{document}
\maketitle
\begin{abstract}
In this paper, we investigate the convergence of language models (LMs) trained under different random seeds, measuring convergence as the expected per-token Kullback--Leibler (KL) divergence across seeds.
By comparing LM convergence as a function of model size and training checkpoint, we identify a four-phase convergence pattern: 
(i) an initial \defn{uniform phase},
(ii) a \defn{sharp-convergence phase},
(iii) a \defn{sharp-divergence phase}, and 
(iv) a \defn{slow-reconvergence phase}.
Further, we observe that larger models reconverge faster in later training stages, while smaller models never actually reconverge; these results suggest that a certain model size may be necessary to learn stable distributions.
Restricting our analysis to specific token frequencies or part-of-speech (PoS) tags further reveals that convergence is uneven across linguistic categories: frequent tokens and function words converge faster and more reliably than their counterparts (infrequent tokens and content words). 
Overall, our findings highlight factors that influence the stability of the learned distributions in model training.
\end{abstract}

\section{Introduction} \label{sec:intro}

At their core, \textbf{language models} (LMs) are distributions over strings, $\ptheta(\subwords)$, trained to approximate a \defn{data-generating distribution} $\ptrue(\subwords)$.
Their massive improvements in recent years---typically attributed to increasing data, compute, and architecture size \citep{kaplan2020scalinglaws,henighan2020scalinglaws}---suggests that LMs are getting ever more similar to this data-generating distribution. 
Notably, if LMs could perfectly fit this data-generating distribution $\ptrue$, they would all converge to the same $\ptheta$.\footnote{See \citet{huh2024platonicrepresentationhypothesis} for an even stronger claim: models not only converge in distribution, but all models across modalities will converge to true ``platonic representations''.}\looseness=-1

\begin{figure}
    \centering
    \begin{subfigure}[b]{\columnwidth}
        \centering
        \includegraphics[trim={.2cm 1cm 0 .1cm},clip,width=\columnwidth,height=3.65cm]{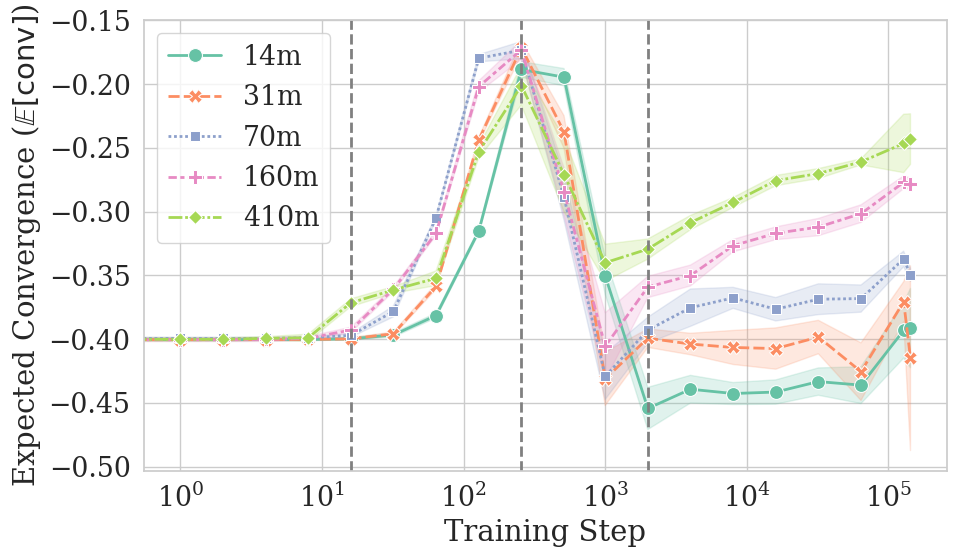}
    \end{subfigure}
    \begin{subfigure}[b]{\columnwidth}
        \centering
        \includegraphics[trim={0 1cm 0 0},clip,width=\columnwidth]{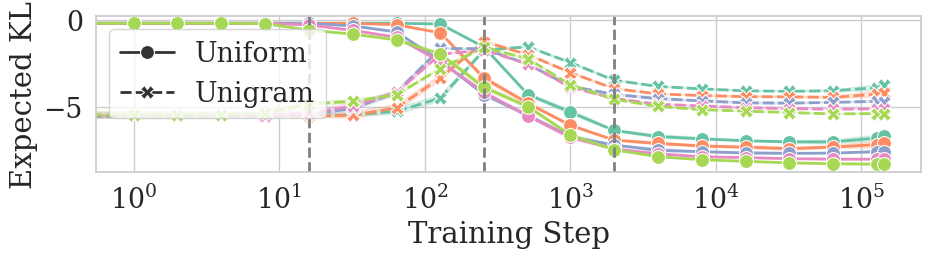}
    \end{subfigure}
    \begin{subfigure}[b]{\columnwidth}
        \centering
        \includegraphics[trim={0 1cm 0 0},clip,width=\columnwidth]{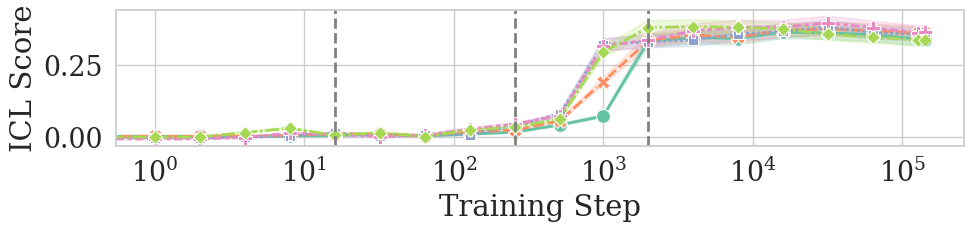}
    \end{subfigure}
    \begin{subfigure}[b]{\columnwidth}
        \centering
        \includegraphics[trim={0 1cm 0 0},clip,width=\columnwidth]{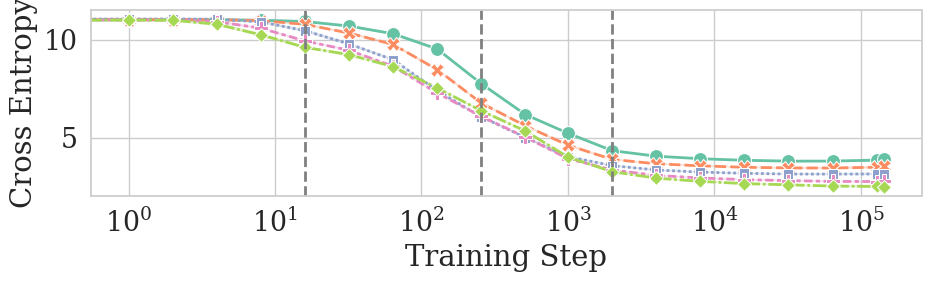}
    \end{subfigure}
    \vspace{-20pt}
    \caption[Estimated $\expconvergence$ across training steps ($x$-axis). Shaded areas represent $1\sigma$ confidence intervals.]{Estimated $\expconvergence$ across training steps ($x$-axis). Shaded areas represent $1\sigma$ confidence intervals.\footnotemark}
    \label{fig:seeds-no-filters}
    \vspace{-10pt}
\end{figure}

In practice, however, this convergence might: (i) not happen uniformly for all contexts; (ii) not happen at all for some contexts.
This is the focus of our study: the convergence (and potential divergence) of LMs across scales, training, and contexts.\footnotetext{
ICL scores were measured as the expected difference in surprisal of a sentence's 500$^\mathrm{th}$ token when: conditioned on the preceding 50 tokens of context vs.\ on 400 tokens.}

Given the scientific and engineering import of language models' training dynamics, a large body of work has examined it \citep[\textit{inter alia}]{saphra-lopez-2019-understanding,wei2022emergent,chen2024sudden,wal2025polypythias}.
We highlight two important previous findings here.
First, early in training, LMs reach a \defn{unigram-output stage}, outputting a mostly context-agnostic distribution which matches word frequencies; only afterward, they start leveraging context \citep{chang-bergen-2022-word,chang-etal-2024-characterizing,learnstatistics}.
Second, and also early in training, transformers go through \defn{induction-head formation}, which enables in-context learning \citep{olsson2022context,tigges2024llm}.

Prior work has also shown that different aspects of input are learned at different rates.
Lexical learning studies, for instance, show that words with different PoS are acquired at different rates \citep{chang-bergen-2022-word,ficarra2025distributional}, suggesting token-specific convergence dynamics.
Relatedly, \citet{evanson-etal-2023-language} show that sentences with more complex structures are learned slower.

As mentioned above, we wish to analyse the convergence of language models here.
To this end, we first define LMs' \defn{convergence} as the negative expected Kullback-Leibler divergence of a model when trained under different seeds.
Relying on this metric, we empirically find that larger models do not simply achieve stronger final convergence, but that convergence happens faster on them.
However, we see that convergence is not monotonic throughout training (see \cref{fig:seeds-no-filters}, top).
After a short initial \defn{uniform phase}, there is a \defn{sharp-convergence phase}; interestingly, this convergence phase coincides with the unigram-output stage found by prior work (\cref{fig:seeds-no-filters}, mid-top).
Afterwards, models follow a \defn{sharp-divergence phase}, where they start learning to use context.
Finally, we see a \defn{slow-reconvergence phase}, in which model predictions seem to stabilise and (at least for larger models) slowly reconverge to a unified solution; interestingly, the transition to this final phase seems to coincide with induction-head formation  (\cref{fig:seeds-no-filters}, mid-bottom).
Notably, these four phases happen while the models monotonically improve (\cref{fig:seeds-no-filters}, bottom); multiple seeds of the same model $\ptheta$ may thus get less similar to each other while simultaneously becoming more similar to the target distribution $\ptrue$.\looseness=-1

Additionally, as in \citet{chang-etal-2024-characterizing}, we study how convergence differs depending on the frequency, part of speech, or final surprisal of a predicted token.
To this end, we define \defn{LM conditional convergence} similar to LM convergence, but conditioning the expectation on a feature of the text (e.g., the target word being a noun). 
Using this metric, our analyses show that, while models' outputs seem to converge when predicting frequent or function words, their final-step convergence on other tokens may be worse than at initialisation.

\section{Convergence and Divergence}
\label{sec:convergence}

In our study, we will measure convergence by analysing whether different models output similar probability distributions.
To that end, we first assume there exists a distribution over model parameters $\ptrue(\btheta)$, induced by a choice of architecture and the optimisation process.
In other words, $\ptrue(\btheta)$ represents a distribution over models trained under different random seeds.
Given this distribution, we define convergence as:
\begin{defin} \label{defn:conv}
    We quantify \textbf{convergence} in context $\context$ as the negative expected divergence between two models $\btheta$ and $\bthetaprime$ sampled from this distribution:
    \begin{align} \label{eq:convergence}
        \convergencefunc(\context) = \expect_{\btheta, \bthetaprime} \bigg[- \distfunc{\context}{\ptheta}{\pthetaprime}\bigg]
    \end{align}
\end{defin}

In theory, we could use any divergence function as $\distfuncbase$.
Here, we  will measure it as the Kullback-Leibler (KL) divergence:\footnote{
Prior work quantifies the convergence of LMs by computing the correlation across seeds in models' predictions \citep[operationalised as, e.g., per-token surprisal, or a downstream task's outputs;][]{chang-etal-2024-characterizing,wal2025polypythias}.
We believe these correlations may hide nuances which $\expconvergence$ captures. 
E.g., two randomly initialised models (which output noisy uniform distributions), output per-token surprisals with near-zero correlations, independently of how close to uniform both their distributions are.
Relatedly, two unigram language models which differ only in their temperature would output surprisals with near-one correlations.}
\begin{align}
    \!\distfunc{\context\!}{\ptheta}{\pthetaprime} 
    &= \kl\Big(\ptheta(\cdot \mid \context) \mid\mid \pthetaprime(\cdot \mid \context) \Big) \!
    \\
    &= \!\sum_{\subword \in \vocab} \ptheta(\subword \mid \context) \log \frac{\ptheta(\subword \mid \context)}{\pthetaprime(\subword \mid \context)} \nonumber
\end{align}
An increase in $\convergencefunc(\context)$ thus indicates convergence, while a decrease in this value indicates divergence.
We chose the KL as it is a standard measure for comparing probability distributions.
In practice, analysing LM convergence for each specific token--context pair can be challenging, and we thus define a global measure of convergence using its expectation.

\begin{defin} \label{defn:expect_conv}
    We quantify \textbf{expected convergence} as the expectation of convergence across contexts:
    \begin{align}
        \expconvergence = \expect_{\context} \bigg[\convergencefunc(\context)\bigg]
    \end{align}
\end{defin}

Notably, while expected convergence gives an overall notion of how convergence behaves across a dataset, it can hide variations in convergence depending on the context and target token. 
To address this, we take inspiration from \citeposs{chang-etal-2024-characterizing} analyses, defining conditional convergence. 
Conditional convergence measures a model's expected convergence conditioned on a specific property.

\begin{defin} \label{defn:cond_conv}
    Let $\vocab_t$ be a set of tokens which have a specific target property to be conditioned on.
    We quantify a model's \textbf{conditional convergence} as the expectation of convergence across tokens and contexts which have this property:%
    \begin{align}
        \condexpconvergence{\vocab_t} = \expect_{\contextplusword} \bigg[\convergencefunc(\context) \mid \subword_t \in \vocab_t \bigg]
    \end{align}
\end{defin}

By computing conditional convergence for different token categories (e.g., nouns, verbs, function words), we can analyse how convergence varies across different tokens.

\section{Experimental Setup}

We now present the main choices made for our experiments.
See \cref{app:detailed_setup} for more details.

\paragraph{Model.}
We analyse language models from the (Poly)Pythia suite here \citep{biderman2023pythia,wal2025polypythias}.\footnote{We also present similar results for MultiBERT in \cref{app:extra_results}.}
For each model size, these LMs' architecture and optimisation procedure induce a distribution $\ptheta(\btheta)$ over parameters, which we use in our definition of convergence (in \cref{defn:conv}).
This suite contains a set of 10 independently trained models per model size, simulating the set of samples $\btheta, \btheta' \sim \ptrue(\btheta)$ we need to estimate $\convergencefunc$ (in \cref{eq:convergence}).
Furthermore, we analyse models with $\{14m, 31m, 70m, 160m, 410m\}$ parameters---the sizes available in PolyPythia---at logarithmically-spaced training steps: $\{0,\allowdisplaybreaks 1,\allowdisplaybreaks 2,\allowdisplaybreaks 4,\allowdisplaybreaks 8, ...,\allowdisplaybreaks 512,\allowdisplaybreaks 1k,\allowdisplaybreaks 2k,\allowdisplaybreaks 4k, ...,\allowdisplaybreaks 128k\}$.
Due to computational restrictions, we selected seeds $1,3,5,7,9$ for our analyses (ignoring other seeds).

\paragraph{Data.}
For our analyses, we use a subset of the Pile's validation set \citep{pile} covering $4{,}662$ tokens; these tokens' contexts form a dataset: $\dataset = \{\subwords_{<t}^{(n)}\}_{n=1}^{N}$.
We assume this data is sampled from the data-generating distribution $\subwords_{<t}^{(n)} \sim \ptrue(\context)$, allowing us to compute an unbiased estimate of expected convergence (in \cref{defn:expect_conv}).

\paragraph{Conditioning Properties.}
Finally, we also analyse how models converge while controlling for three properties: a token $\subword_t$'s frequency, its part-of-speech (PoS), and its final-surprisal.
We estimate tokens' frequencies by counting them on the Pile's validation set.
We estimate PoS using the NLTK part-of-speech tagger \citep{nltk}.\footnote{We detail how we convert from word- (output by NLTK) to subword-level tags (Pythia's tokens) in \cref{app:detailed_setup}.}
We compute final-surprisal by---for a specific model size---using its last checkpoint to compute each token's surprisal: $- \log \ptheta(\subword_t \mid \context)$.
To estimate conditional convergence (\cref{defn:cond_conv}), we then define $\vocab_t$ using either log-spaced frequency or final-surprisals bins, or PoS classes.

\section{Four Phases of Expected Convergence}

\cref{fig:seeds-no-filters} (top) presents our estimates of the expected convergence, i.e., $\expconvergence$, for models of different sizes and across training steps.
In this figure, we see that convergence progresses across training in four clearly distinct phases.

\paragraph{Uniform Phase.}
This initial phase is roughly observed until step 16 and reflects a shared starting point of training, with models of different sizes presenting similar convergences.
This has a simple explanation.
As shown in \cref{fig:seeds-no-filters} (mid-top), all LMs' outputs start similar to the uniform distribution, which is enforced by their parameters' initialisation.
Interestingly, there is almost no change in convergence during this phase, which may be explained by the small learning rates used at these steps.
(For convenience, we present Pythias' learning rates across training in \cref{fig:learning_rates} in \cref{app:learning_rates}).

\paragraph{Sharp-convergence Phase.}
This second phase is roughly observed between steps 16 and 256, being characterised by a sharp increase in model similarity.
As can be seen in \cref{fig:seeds-no-filters} (mid-top), this phase corresponds quite clearly to a shift in LMs from mimicking a uniform to a unigram distribution.\footnote{As discussed in \cref{sec:intro}, \citet{chang-bergen-2022-word} originally reported this unigram-output stage of LM learning.}
As shown by prior work, these are also the training steps at which models' predictions (either in terms of surprisal, or downstream tasks outputs) seem to have maximal correlations with one another \citep{chang-etal-2024-characterizing,wal2025polypythias}.

\paragraph{Sharp-divergence Phase.}
This third phase of training occurs between steps 256 and 2k, being characterised by a sharp decrease in model similarity.
Interestingly, this phase coincides with the moment LMs start diverging from the unigram distribution, implying that, at least initially, LMs use of context differs significantly across seeds.
Notably, the cross-entropy of Pythia models decreases monotonically throughout training (see \cref{fig:seeds-no-filters}, bottom), making such a sharp-divergence phase surprising: 
these LMs seem to monotonically approximate $\ptrue$, but each does so in a different way.

\begin{figure*}[t]
    \centering
    \begin{subfigure}[b]{0.48\linewidth}
        \centering
        \includegraphics[width=\columnwidth]{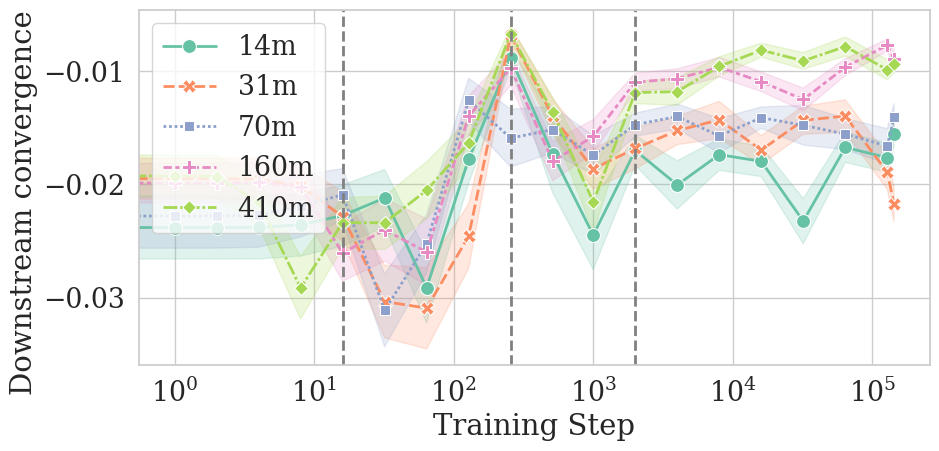}
    \end{subfigure}%
    \hfill
    \begin{subfigure}[b]{0.48\linewidth}
        \centering
        \includegraphics[width=\columnwidth]{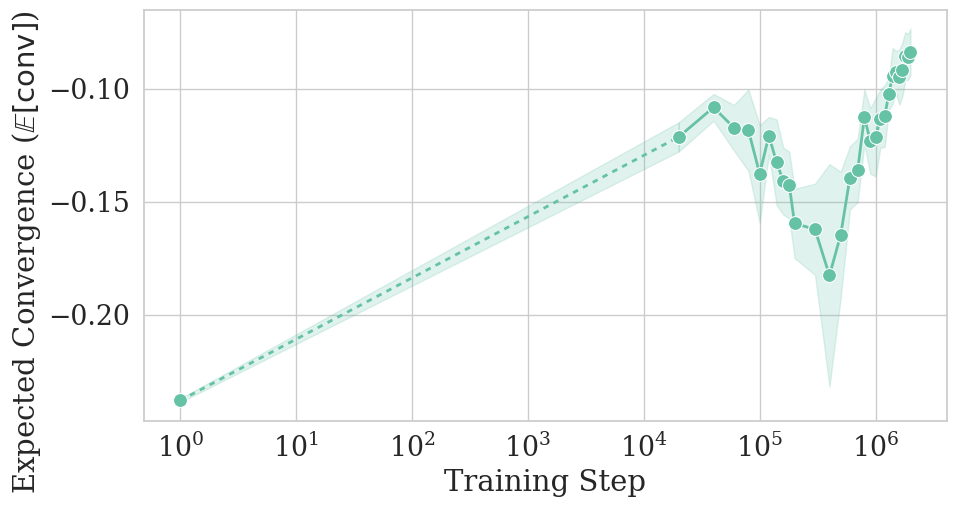}
    \end{subfigure}%
    \vspace{-5pt}
    \caption{Estimated $\expconvergence$ across training steps: (left) on the Pythia model suite on BLiMP with $1\bar{\sigma}$ confidence intervals, (right) on the MultiBERT model suite on masked language modelling with $1\sigma$ confidence intervals.}
    \label{fig:blimp}
\end{figure*}

\paragraph{Slow-reconvergence Phase.}
This final phase starts around step 2k, and is characterised by a slow increase in model similarity.
Interestingly, as can be seen in \cref{fig:seeds-no-filters} (mid-bottom),
the step at which this phase starts coincides with an increase in in-context learning (ICL) scores \citep{olsson2022context} and thus to induction-heads formation.
This suggests that induction heads may not only enable in-context learning in large models, but also stabilise the training of transformer-based LMs.
Further, the steepness of this final downward trend depends on model size, with larger models converging faster than smaller ones.
In fact, for the smallest models, model convergence seems to mostly stabilise at this point, and they end training with similar $\expconvergence$ to what they begin with, implying that these models do not in fact converge to a shared solution.

\subsection{Convergence on Other Tasks and Models}
\label{app:extra_results}

\newcommand{\phat}{\widehat{\ptheta}}
\newcommand{\phatprime}{\widehat{\ptheta}'}
\newcommand{\subwordsgrammatical}{\subwords_{\text{\cmark}}}
\newcommand{\subwordsungrammatical}{\subwords_{\text{\xmark}}}

We now assess whether model convergence dynamics are similar in: (i) downstream tasks, conducting experiments on BLiMP \cite{blimp}; (ii) other models, conducting experiments with the MultiBERT model suite \cite{multiberts}.

\paragraph{Convergence on Downstream Tasks.}
BLiMP is a dataset composed of pairs of grammatical and ungrammatical sentences, $\dataset = \{\subwordsgrammatical^{(n)}, \subwordsungrammatical^{(n)}\}_{n=1}^{N}$, and whose task is to identify the grammatical one.
For each of these pairs, models are then evaluated on whether they place more probability on $\subwordsgrammatical$ than on $\subwordsungrammatical$.
In this task, we thus restrict the support of our models' distribution to these two sentences:
\begin{align}
    &\phat(\subwordsgrammatical) = \scaleto{\frac{\ptheta(\subwordsgrammatical)}{\ptheta(\subwordsgrammatical) + \ptheta(\subwordsungrammatical)}}{20pt}, \quad 
    \phat(\subwordsungrammatical) = \scaleto{\frac{\ptheta(\subwordsungrammatical)}{\ptheta(\subwordsgrammatical) + \ptheta(\subwordsungrammatical)}}{20pt} \nonumber
\end{align}
We then use this limited-support distribution to compute models' convergence as in \cref{sec:convergence}, but with:
\begin{align}
    &\distfuncbase_{n}(\ptheta, \pthetaprime)
    = \sum_{\subwords \in \{\subwordsgrammatical^{(n)}, \subwordsungrammatical^{(n)}\}}\!\! \phat(\subwords) \log \frac{\phat(\subwords)}{\phatprime(\subwords)}
\end{align}
We present these downstream convergence results in \cref{fig:blimp} (left). 
This figure reveals that the downstream convergence pattern broadly mirrors the phases observed before, supporting the hypothesis that these training dynamics manifest at the task-level as well.

\paragraph{Convergence on Masked LMs.}
The MultiBERT model suite is composed of bidirectional transformers---as opposed to Pythia's autoregressive models---and allows us to evaluate whether convergence dynamics are similar in such masked language models.
Unfortunately, the available checkpoints do not include steps between 0 and 20k, which prevents us from observing (i) an initial \defn{uniform phase} or (ii) a \defn{sharp-convergence phase}.
However, the remaining dynamics appear consistent with the results on Pythia, including: (iii) a \defn{sharp-divergence phase} followed by (iv) a \defn{slow-reconvergence phase} (see \cref{fig:blimp}, right).

\begin{figure*}
    \begin{subfigure}[b]{0.33\linewidth}
        \centering
        \includegraphics[trim={0cm 0 0cm 0},clip,width=\linewidth]{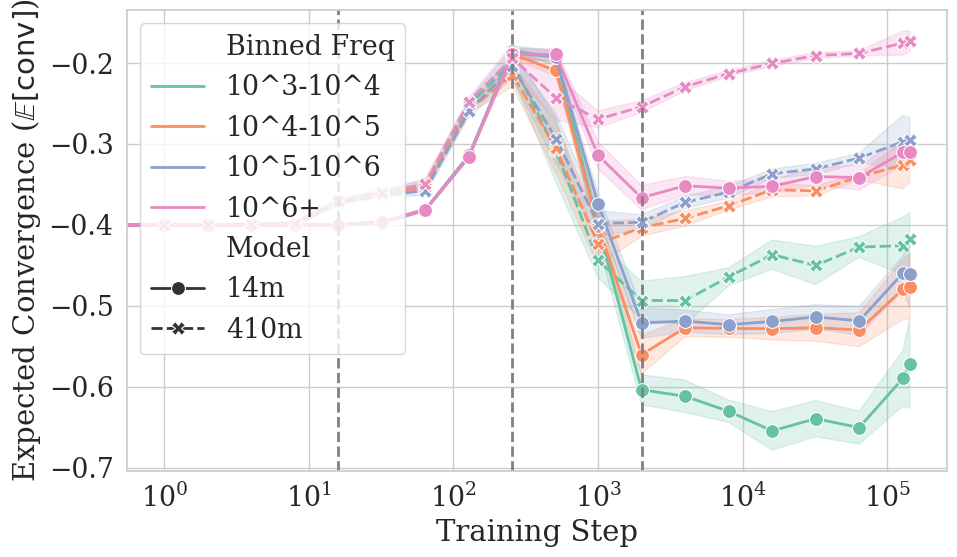}
    \end{subfigure}%
    \begin{subfigure}[b]{0.33\linewidth}
        \centering
        \includegraphics[trim={0cm 0 0cm 0},clip,width=\linewidth]{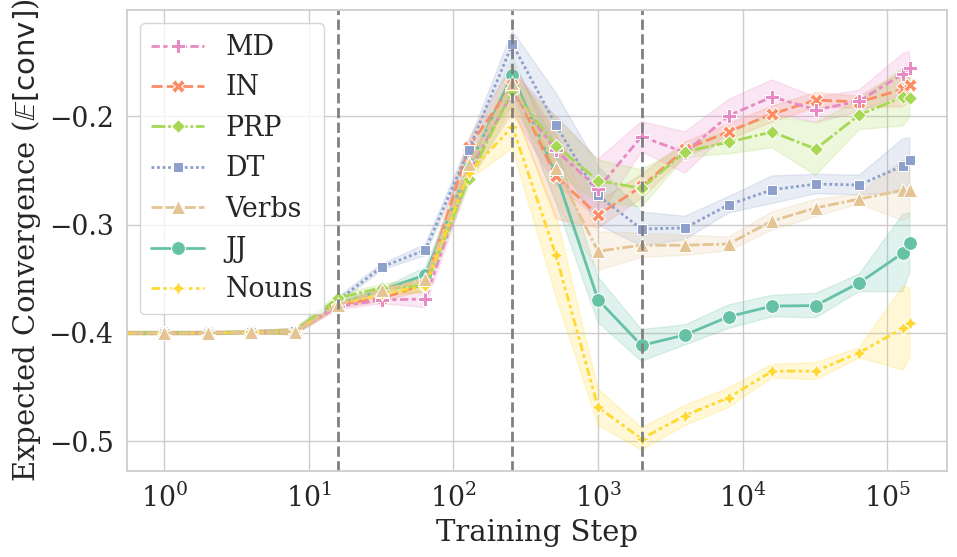}
    \end{subfigure}%
    \begin{subfigure}[b]{0.33\linewidth}
        \centering
        \includegraphics[trim={0cm 0 0cm 0},clip,width=\linewidth]{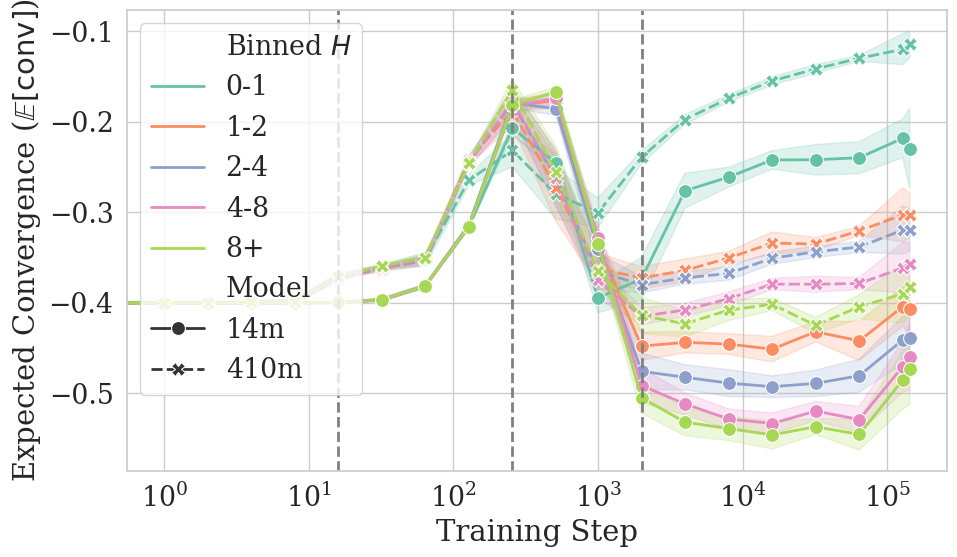}
    \end{subfigure}%
    \vspace{-5pt}
    \caption{$\condexpconvergence{\vocab_t}$ of selected models with $1\sigma$ confidence intervals. Conditioning property: (left) frequency; (center) parts of speech; (right) final surprisal.}
    \label{fig:seeds_cond_conv}
\end{figure*}

\section{Conditional Convergence}

We now analyse conditional convergences: i.e., how convergence changes as a function of different contextual properties of a token.
These results are presented in \cref{fig:seeds_cond_conv}, where conditional convergences are presented for tokens based on either frequency, PoS, or final-surprisal.\footnote{We note that, within any of these categories---similarly to the general case-- LMs present (almost) monotonically decreasing cross-entropy curves (see \cref{fig:cond_cross_entropies} in \cref{app:cond_cross_entropies}).}
Notably, for all these conditioning properties, the two initial phases of training (the uniform and sharp-convergence phases) present similar trends.
This is likely because until the third phase of training (the sharp-divergence phase), LMs are not using context to make predictions.
We will thus focus on the third and fourth convergence phases here.

\paragraph{Frequency.}
\cref{fig:seeds_cond_conv} (left) presents the conditional convergence $\condexpconvergence{\vocab_t}$ for tokens in varying frequency bins.
During the third training phase, frequent tokens' convergence stabilises at a relatively high value.
In contrast, infrequent tokens fully suffer from a sharp-divergence, highlighting their greater variability and sensitivity to random seed.
Interestingly, final convergence in infrequent tokens is smaller than initial convergence, suggesting LMs diverge on these tokens across training.

\paragraph{PoS Tags.}
\cref{fig:seeds_cond_conv} (center) presents the conditional convergence $\condexpconvergence{\vocab_t}$ for tokens with varying PoS tags.
This figure shows that content words (nouns, adjectives (JJ) and verbs) diverge more once entering the sharp-divergence phase than function words
(determiners (DT), personal pronouns (PRP), prepositions or subordinating conjunctions (IN), and modal auxiliary words (MD)).
Furthermore, function words achieve higher final convergence, whereas content words are more divergent.%
\footnote{See \cref{app:cond_conv__nouns_and_verbs} for a detailed analysis of how convergence changes across subclasses of nouns and verbs. Further, as different conditioning properties may be correlated (function words are typically frequent), we present a linear regression analysis of $\convergencefunc(\context)$ in \cref{app:linear_regression} to jointly analyse these properties impact; this analysis supported our main results here.}

\paragraph{Final Surprisal.}
\cref{fig:seeds_cond_conv} (right) presents the conditional convergence $\condexpconvergence{\vocab_t}$ for tokens within varying final-surprisal bins.
This figure reveals that a token's convergence may be fairly different depending on how predictable it is.
In particular, this figure shows that tokens with very low final-surprisal show strong convergence by the end of training.
Convergence behaviour across tokens with higher final-surprisals (from 1 to 8+ bits), however, seems quite similar, with final-surprisal thus not greatly impacting convergence for these tokens.

\subsection{Variance in Convergence across Tokens}

\begin{figure}[t]
    \centering
    \includegraphics[width=\linewidth]{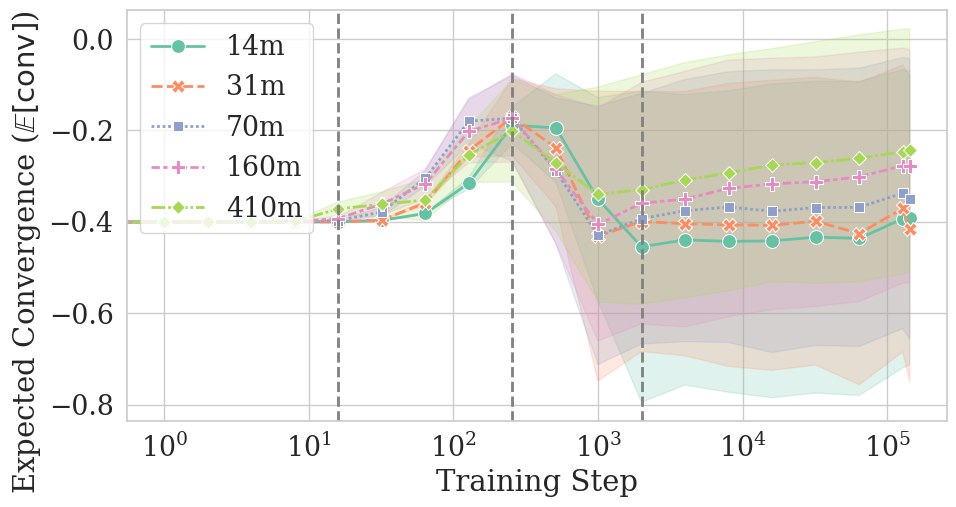}
    \vspace{-16pt}
    \caption{$\convergencefunc(\context)$ across training, with shaded areas representing its standard across contexts $\context$.}
    \label{fig:seeds_context_variance}
\end{figure}

Finally, we also analyse the variance of $\convergencefunc(\context)$ across contexts $\context$ throughout training.
\Cref{fig:seeds_context_variance} presents these results.
Interestingly, this figure shows that the initial two phases have very little variance in convergence across contexts $\context$.
The final two phases, however, present a significant increase in this variance.
This again highlights the non-contextual nature of the two initial convergence phases.

\section{Conclusion}

Our analyses reveal that convergence in language models is far from uniform. 
While global metrics like cross-entropy steadily improve, they conceal substantial variation in how individual tokens and contexts behave across training.
We find that, across training, LMs' convergence goes through four phases, each with distinct characteristics.\footnote{
\citet{li2025tracingrepresentationgeometrylanguage} shows that the effective rank of LMs' representations presents similarly distinct phases across training.}
Additionally, we find that token frequency plays a dominant role in convergence: frequent tokens converge quickly and consistently, whereas rare tokens often diverge. 
Similarly, convergence is strongly shaped by linguistic features: function words exhibit stable predictions, while content words remain volatile (a result reminiscent of \citealp{chang-bergen-2022-word}).
Finally, we find that larger models tend to converge more consistently.

\section*{Limitations}

Our analysis is limited in several ways. 
First, our analysis of a $\convergencefunc(\context)$ measure based on the token-level KL prevents us from comparing different model families that rely on different tokenisers, as different distribution supports in $\ptheta(\subword_t \mid \context)$ would prevent us from calculating the KL divergence.
This could potentially be mitigated by converting these distributions to the byte- or word-level \citep{pimentel-meister-2024-compute,phan2025exact}, but we leave that for future work.
Second, due to computational constraints, our experiments were conducted on a small subset of the Pile's validation set.
Third, our analysis is restricted to English-language data, leaving open questions about whether similar convergence dynamics occur in other languages and in multilingual settings. 
Finally, the presence of learning rate warm-up phases during early training, where we observe the most rapid shifts in model behaviour, may introduce artifacts that affect our interpretation of convergence and divergence.

\section*{Acknowledgements}

We would like to thank Pietro Lesci for generously providing us with helpful references and feedback on earlier versions of this work. The authors also thank Prof.\ Yoon Kim for his valuable input and for providing computational resources. We thank Prof. Thomas Hofmann for supporting this research in his lab and for financing the conference participation, as well as for his broader guidance during the thesis work. Kyle Mahowald acknowledges funding from NSF CAREER grant 2339729.

\bibliography{custom}

\begin{thebibliography}{22}
\providecommand{\natexlab}[1]{#1}

\bibitem[{Belrose et~al.(2024)Belrose, Pope, Quirke, Mallen, and
  Fern}]{learnstatistics}
Nora Belrose, Quintin Pope, Lucia Quirke, Alex Mallen, and Xiaoli Fern. 2024.
\newblock \href {https://arxiv.org/abs/2402.04362} {Neural networks learn
  statistics of increasing complexity}.
\newblock \emph{Preprint}, arXiv:2402.04362.

\bibitem[{Biderman et~al.(2023)Biderman, Schoelkopf, Anthony, Bradley,
  O’Brien, Hallahan, Khan, Purohit, Prashanth, Raff
  et~al.}]{biderman2023pythia}
Stella Biderman, Hailey Schoelkopf, Quentin~Gregory Anthony, Herbie Bradley,
  Kyle O’Brien, Eric Hallahan, Mohammad~Aflah Khan, Shivanshu Purohit,
  USVSN~Sai Prashanth, Edward Raff, et~al. 2023.
\newblock \href {https://arxiv.org/abs/2304.01373} {{P}ythia: {A} suite for
  analyzing large language models across training and scaling}.
\newblock In \emph{International Conference on Machine Learning}, pages
  2397--2430. PMLR.

\bibitem[{Bird and Loper(2004)}]{nltk}
Steven Bird and Edward Loper. 2004.
\newblock \href {https://aclanthology.org/P04-3031/} {{NLTK}: The natural
  language toolkit}.
\newblock In \emph{Proceedings of the {ACL} Interactive Poster and
  Demonstration Sessions}, pages 214--217, Barcelona, Spain. Association for
  Computational Linguistics.

\bibitem[{Chang and Bergen(2022)}]{chang-bergen-2022-word}
Tyler~A. Chang and Benjamin~K. Bergen. 2022.
\newblock \href {https://doi.org/10.1162/tacl_a_00444} {Word acquisition in
  neural language models}.
\newblock \emph{Transactions of the Association for Computational Linguistics},
  10:1--16.

\bibitem[{Chang et~al.(2024)Chang, Tu, and
  Bergen}]{chang-etal-2024-characterizing}
Tyler~A. Chang, Zhuowen Tu, and Benjamin~K. Bergen. 2024.
\newblock \href {https://doi.org/10.1162/tacl_a_00708} {Characterizing learning
  curves during language model pre-training: Learning, forgetting, and
  stability}.
\newblock \emph{Transactions of the Association for Computational Linguistics},
  12:1346--1362.

\bibitem[{Chen et~al.(2024)Chen, Shwartz-Ziv, Cho, Leavitt, and
  Saphra}]{chen2024sudden}
Angelica Chen, Ravid Shwartz-Ziv, Kyunghyun Cho, Matthew~L. Leavitt, and Naomi
  Saphra. 2024.
\newblock \href {https://openreview.net/forum?id=MO5PiKHELW} {Sudden drops in
  the loss: Syntax acquisition, phase transitions, and simplicity bias in
  {MLM}s}.
\newblock In \emph{The Twelfth International Conference on Learning
  Representations}.

\bibitem[{Evanson et~al.(2023)Evanson, Lakretz, and
  King}]{evanson-etal-2023-language}
Linnea Evanson, Yair Lakretz, and Jean~R{\'e}mi King. 2023.
\newblock \href {https://doi.org/10.18653/v1/2023.findings-acl.773} {Language
  acquisition: do children and language models follow similar learning stages?}
\newblock In \emph{Findings of the Association for Computational Linguistics:
  ACL 2023}, pages 12205--12218, Toronto, Canada. Association for Computational
  Linguistics.

\bibitem[{Ficarra et~al.(2025)Ficarra, Cotterell, and
  Warstadt}]{ficarra2025distributional}
Filippo Ficarra, Ryan Cotterell, and Alex Warstadt. 2025.
\newblock \href {https://arxiv.org/abs/2502.05892} {A distributional
  perspective on word learning in neural language models}.
\newblock \emph{Preprint}, arXiv:2502.05892.

\bibitem[{Gao et~al.(2020)Gao, Biderman, Black, Golding, Hoppe, Foster, Phang,
  He, Thite, Nabeshima, Presser, and Leahy}]{pile}
Leo Gao, Stella Biderman, Sid Black, Laurence Golding, Travis Hoppe, Charles
  Foster, Jason Phang, Horace He, Anish Thite, Noa Nabeshima, Shawn Presser,
  and Connor Leahy. 2020.
\newblock \href {https://arxiv.org/abs/2101.00027} {The {P}ile: {A}n 800{GB}
  dataset of diverse text for language modeling}.
\newblock \emph{Preprint}, arXiv:2101.00027.

\bibitem[{Henighan et~al.(2020)Henighan, Kaplan, Katz, Chen, Hesse, Jackson,
  Jun, Brown, Dhariwal, Gray, Hallacy, Mann, Radford, Ramesh, Ryder, Ziegler,
  Schulman, Amodei, and McCandlish}]{henighan2020scalinglaws}
Tom Henighan, Jared Kaplan, Mor Katz, Mark Chen, Christopher Hesse, Jacob
  Jackson, Heewoo Jun, Tom~B. Brown, Prafulla Dhariwal, Scott Gray, Chris
  Hallacy, Benjamin Mann, Alec Radford, Aditya Ramesh, Nick Ryder, Daniel~M.
  Ziegler, John Schulman, Dario Amodei, and Sam McCandlish. 2020.
\newblock \href {https://arxiv.org/abs/2010.14701} {Scaling laws for
  autoregressive generative modeling}.
\newblock \emph{Preprint}, arXiv:2010.14701.

\bibitem[{Huh et~al.(2024)Huh, Cheung, Wang, and
  Isola}]{huh2024platonicrepresentationhypothesis}
Minyoung Huh, Brian Cheung, Tongzhou Wang, and Phillip Isola. 2024.
\newblock \href {https://arxiv.org/abs/2405.07987} {The platonic representation
  hypothesis}.
\newblock \emph{Preprint}, arXiv:2405.07987.

\bibitem[{Kaplan et~al.(2020)Kaplan, McCandlish, Henighan, Brown, Chess, Child,
  Gray, Radford, Wu, and Amodei}]{kaplan2020scalinglaws}
Jared Kaplan, Sam McCandlish, Tom Henighan, Tom~B. Brown, Benjamin Chess, Rewon
  Child, Scott Gray, Alec Radford, Jeffrey Wu, and Dario Amodei. 2020.
\newblock \href {https://arxiv.org/abs/2001.08361} {Scaling laws for neural
  language models}.
\newblock \emph{Preprint}, arXiv:2001.08361.

\bibitem[{Li et~al.(2025)Li, Agrawal, Ghosh, Teru, Santoro, Lajoie, and
  Richards}]{li2025tracingrepresentationgeometrylanguage}
Melody~Zixuan Li, Kumar~Krishna Agrawal, Arna Ghosh, Komal~Kumar Teru, Adam
  Santoro, Guillaume Lajoie, and Blake~A. Richards. 2025.
\newblock \href {https://arxiv.org/abs/2509.23024} {Tracing the representation
  geometry of language models from pretraining to post-training}.
\newblock \emph{Preprint}, arXiv:2509.23024.

\bibitem[{Olsson et~al.(2022)Olsson, Elhage, Nanda, Joseph, DasSarma, Henighan,
  Mann, Askell, Bai, Chen, Conerly, Drain, Ganguli, Hatfield-Dodds, Hernandez,
  Johnston, Jones, Kernion, Lovitt, Ndousse, Amodei, Brown, Clark, Kaplan,
  McCandlish, and Olah}]{olsson2022context}
Catherine Olsson, Nelson Elhage, Neel Nanda, Nicholas Joseph, Nova DasSarma,
  Tom Henighan, Ben Mann, Amanda Askell, Yuntao Bai, Anna Chen, Tom Conerly,
  Dawn Drain, Deep Ganguli, Zac Hatfield-Dodds, Danny Hernandez, Scott
  Johnston, Andy Jones, Jackson Kernion, Liane Lovitt, Kamal Ndousse, Dario
  Amodei, Tom Brown, Jack Clark, Jared Kaplan, Sam McCandlish, and Chris Olah.
  2022.
\newblock \href
  {https://transformer-circuits.pub/2022/in-context-learning-and-induction-heads/index.html}
  {In-context learning and induction heads}.
\newblock \emph{Transformer Circuits Thread}.

\bibitem[{Phan et~al.(2025)Phan, Amos, Gat, Havasi, Muckley, and
  Ullrich}]{phan2025exact}
Buu Phan, Brandon Amos, Itai Gat, Marton Havasi, Matthew~J. Muckley, and Karen
  Ullrich. 2025.
\newblock \href {https://openreview.net/forum?id=zGej22CBnS} {Exact byte-level
  probabilities from tokenized language models for {FIM}-tasks and model
  ensembles}.
\newblock In \emph{The Thirteenth International Conference on Learning
  Representations}.

\bibitem[{Pimentel and Meister(2024)}]{pimentel-meister-2024-compute}
Tiago Pimentel and Clara Meister. 2024.
\newblock \href {https://doi.org/10.18653/v1/2024.emnlp-main.1020} {How to
  compute the probability of a word}.
\newblock In \emph{Proceedings of the 2024 Conference on Empirical Methods in
  Natural Language Processing}, pages 18358--18375, Miami, Florida, USA.
  Association for Computational Linguistics.

\bibitem[{Saphra and Lopez(2019)}]{saphra-lopez-2019-understanding}
Naomi Saphra and Adam Lopez. 2019.
\newblock \href {https://doi.org/10.18653/v1/N19-1329} {Understanding learning
  dynamics of language models with {SVCCA}}.
\newblock In \emph{Proceedings of the 2019 Conference of the North {A}merican
  Chapter of the Association for Computational Linguistics: Human Language
  Technologies, Volume 1 (Long and Short Papers)}, pages 3257--3267,
  Minneapolis, Minnesota. Association for Computational Linguistics.

\bibitem[{Sellam et~al.(2022)Sellam, Yadlowsky, Tenney, Wei, Saphra, D'Amour,
  Linzen, Bastings, Turc, Eisenstein, Das, and Pavlick}]{multiberts}
Thibault Sellam, Steve Yadlowsky, Ian Tenney, Jason Wei, Naomi Saphra,
  Alexander D'Amour, Tal Linzen, Jasmijn Bastings, Iulia~Raluca Turc, Jacob
  Eisenstein, Dipanjan Das, and Ellie Pavlick. 2022.
\newblock \href {https://openreview.net/forum?id=K0E_F0gFDgA} {The
  multi{BERT}s: {BERT} reproductions for robustness analysis}.
\newblock In \emph{International Conference on Learning Representations}.

\bibitem[{Tigges et~al.(2024)Tigges, Hanna, Yu, and Biderman}]{tigges2024llm}
Curt Tigges, Michael Hanna, Qinan Yu, and Stella Biderman. 2024.
\newblock \href {https://openreview.net/forum?id=3Ds5vNudIE} {{LLM} circuit
  analyses are consistent across training and scale}.
\newblock In \emph{The Thirty-eighth Annual Conference on Neural Information
  Processing Systems}.

\bibitem[{van~der Wal et~al.(2025)van~der Wal, Lesci, M{\"u}ller-Eberstein,
  Saphra, Schoelkopf, Zuidema, and Biderman}]{wal2025polypythias}
Oskar van~der Wal, Pietro Lesci, Max M{\"u}ller-Eberstein, Naomi Saphra, Hailey
  Schoelkopf, Willem Zuidema, and Stella Biderman. 2025.
\newblock \href {https://openreview.net/forum?id=bmrYu2Ekdz} {{P}oly{P}ythias:
  {S}tability and outliers across fifty language model pre-training runs}.
\newblock In \emph{The Thirteenth International Conference on Learning
  Representations}.

\bibitem[{Warstadt et~al.(2020)Warstadt, Parrish, Liu, Mohananey, Peng, Wang,
  and Bowman}]{blimp}
Alex Warstadt, Alicia Parrish, Haokun Liu, Anhad Mohananey, Wei Peng, Sheng-Fu
  Wang, and Samuel~R. Bowman. 2020.
\newblock \href {https://doi.org/10.1162/tacl_a_00321} {{BL}i{MP}: The
  benchmark of linguistic minimal pairs for {E}nglish}.
\newblock \emph{Transactions of the Association for Computational Linguistics},
  8:377--392.

\bibitem[{Wei et~al.(2022)Wei, Tay, Bommasani, Raffel, Zoph, Borgeaud,
  Yogatama, Bosma, Zhou, Metzler, Chi, Hashimoto, Vinyals, Liang, Dean, and
  Fedus}]{wei2022emergent}
Jason Wei, Yi~Tay, Rishi Bommasani, Colin Raffel, Barret Zoph, Sebastian
  Borgeaud, Dani Yogatama, Maarten Bosma, Denny Zhou, Donald Metzler, Ed~H.
  Chi, Tatsunori Hashimoto, Oriol Vinyals, Percy Liang, Jeff Dean, and William
  Fedus. 2022.
\newblock \href {https://openreview.net/forum?id=yzkSU5zdwD} {Emergent
  abilities of large language models}.
\newblock \emph{Transactions on Machine Learning Research}.
\newblock Survey Certification.

\end{thebibliography}

\newpage
\appendix

\section{Detailed Experimental Setup}\label{app:detailed_setup}

In this section, we expand on experimental choices used to estimate convergence.

\paragraph{Choice of $\ptrue(\btheta)$ (Or, Analysed Model Architecture and Optimisation Process).}
Our definition of convergence (in \cref{defn:conv}) relies on a distribution over models $\ptrue(\btheta)$, which is itself induced by a choice of model architecture and optimisation process.
Here, we will use the distribution $\ptrue(\btheta)$ induced by Pythia's training process \citep{biderman2023pythia}; see their paper for details.
The Pythia model suite includes model architectures of different sizes, and we analyse here models with $\{14m, 31m, 70m, 160m, 410m\}$ parameters.\footnote{We restrict our analyses to these model sizes, as only those are covered by the PolyPythia model suite, whose relevance we expand on in the next paragraph.}
Further, 154 checkpoints were released for each of these model sizes, allowing us to analyse how convergence evolves across training.
Here, we analyse training steps at logarithmically-spaced intervals; specifically, we analyse checkpoints: $\{0, 1, 2, 4, 8, ..., 512, 1k, 2k, 4k, ..., 128k\}$.

\paragraph{Estimating Convergence (Or, Analysed Model).}
To compute $\convergencefunc(\context)$ we need not only to choose a distribution $\ptrue(\btheta)$, but to take an expectation over it.
This is infeasible for large language models.
We can, however, estimate $\convergencefunc(\context)$ using pairs of independently sampled models $\btheta, \btheta' \sim \ptrue(\btheta)$.
Luckily, \citet{wal2025polypythias} recently presented the PolyPythia model suite, an extension of the original Pythia model suite with multiple trained models---using different randomisation seeds---for each model size.
We treat each pair of models in the PolyPythia suite as a sample $\btheta, \btheta' \sim \ptrue(\btheta)$, which we use to estimate convergence.
Due to computational restrictions, we selected seeds $1,3,5,7,9$ for our analyses (ignoring the other 5 seeds).

\paragraph{Estimating Expected Convergence (Or, Analysed Data).}
To compute $\expconvergence$, we must take an expectation over contexts $\context \sim \ptrue(\context)$, which is again computationally infeasible.
To avoid this issue, we use a data set $\dataset = \{\subwords_{<t}^{(n)}\}_{n=1}^{N}$ of samples which we assume to be drawn from the true distribution $\ptrue(\context)$; this allows us to compute an unbiased estimate of expected convergence.
More specifically, we used samples from the Pile validation set \citep{pile} that covered the $4662$ tokens.\looseness=-1

\paragraph{Choice of $\vocab_t$ (Or, Analysed Token Property).}
Our definition of conditional convergence (\cref{defn:cond_conv}) relies on a choice of token property we which to condition on.
Here, we will consider three such properties: a token $\subword_t$'s frequency, its part-of-speech (PoS), and its final surprisal.
We estimate a tokens' frequency by counting the number of times it appears on our dataset $\dataset$; we then define log-spaced bins $\vocab_t$ which we use to analyse tokens within those frequencies.
We estimate a token's PoS tag using the NLTK part-of-speech tagger \citep{nltk}.
Since our dataset is primarily in English, we use the standard PoS tags for this language.
However, because NLTK and Pythia employ different tokenisation methods, we implement a mapping procedure to align PoS tags with the tokenised outputs.
First, PoS tags are assigned at the word level using NLTK. These tags are then mapped to individual characters in the raw text. 
Finally, after tokenisation, each token inherits the tag that corresponds to the majority of its characters.
Tokens without a majority label or with an "UNK" (unknown) label are excluded from our analysis.
This process is illustrated in \cref{fig:pos-mapping}:

\begin{figure}[h]
    \centering
    \includegraphics[width=\columnwidth]{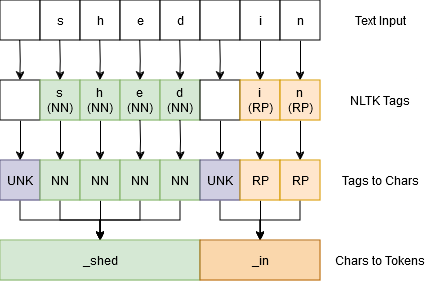}
    \caption{Illustration of the mapping of part of speech tags to tokens.}
    \label{fig:pos-mapping}
\end{figure}

\noindent
Finally, we estimate a tokens' final surprisal by, for each analysed model, using the model's final checkpoint to estimate the token's surprisal; similarly to our frequency analysis, we then define log-spaced bins $\vocab_t$ which we use to analyse tokens within those surprisal ranges.

\newpage

\section{Pythias' Learning Rates} \label{app:learning_rates}

\begin{figure}[h]
    \centering
    \includegraphics[width=\linewidth]{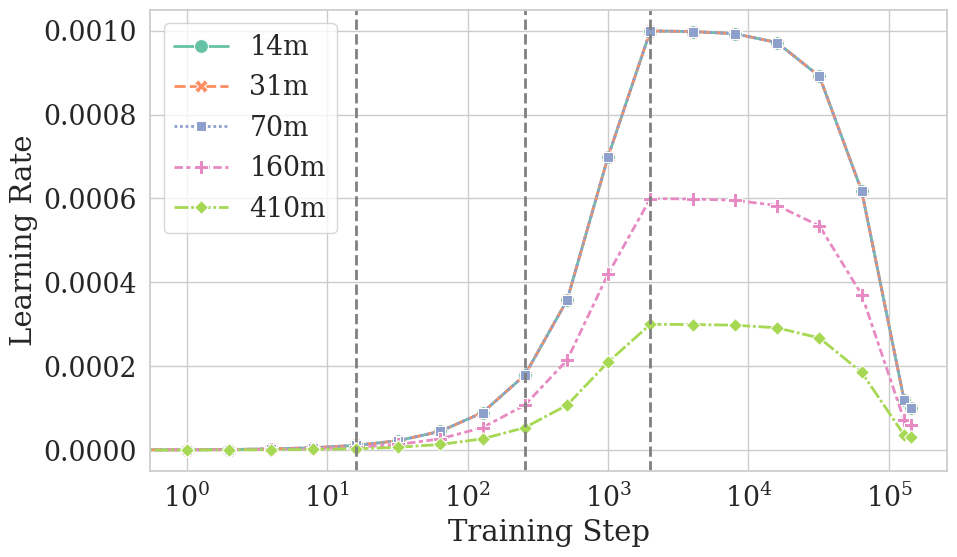}
    \caption{Learning rates for the different Pythia models.}
    \label{fig:learning_rates}
\end{figure}

\section{Conditional Convergence on Nouns and Verbs PoS}
\label{app:cond_conv__nouns_and_verbs}

\begin{figure}[h]
    \centering
    \begin{subfigure}[b]{0.5\linewidth}
        \centering
        \includegraphics[trim={1cm 0 0cm 0},clip,width=\linewidth]{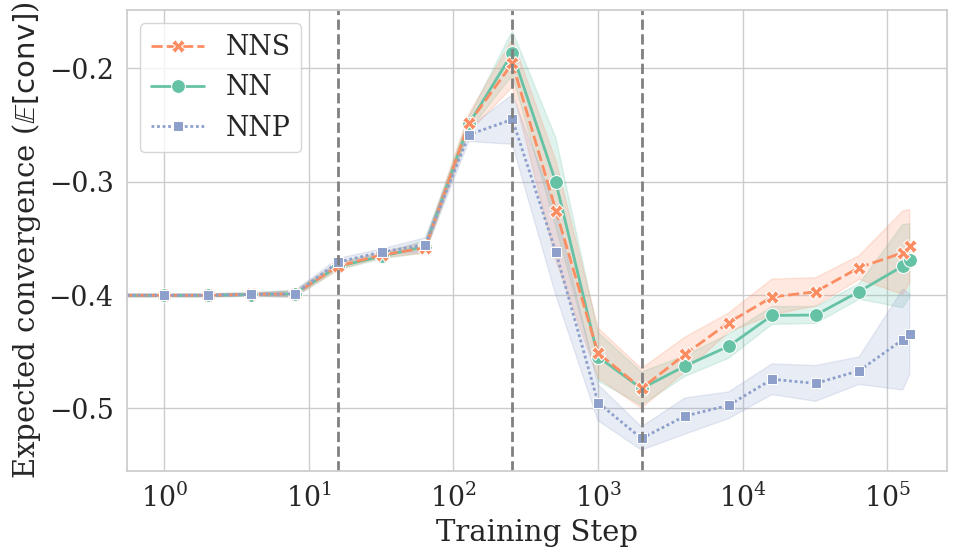}
    \end{subfigure}%
    \begin{subfigure}[b]{0.5\linewidth}
        \centering
        \includegraphics[trim={1cm 0 0cm 0},clip,width=\linewidth]{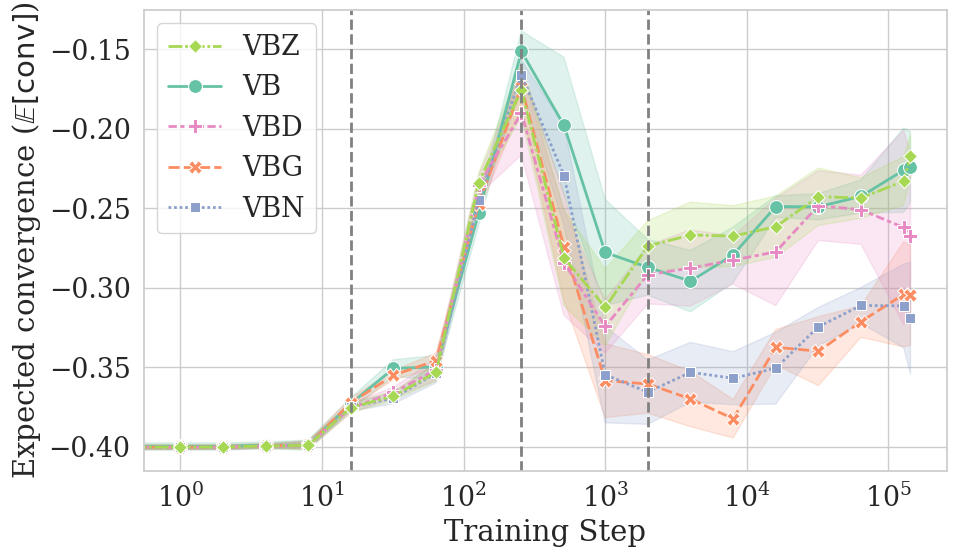}
    \end{subfigure}%
    \caption{$\condexpconvergence{\vocab_t}$ of selected models with $1\sigma$ confidence intervals. Conditioning property: (left) Nouns' PoS; (right) Verbs' PoS.}
    \label{fig:seeds_cond_conv__nouns_and_verbs}
\end{figure}

\paragraph{Differences in Nouns.}
\Cref{fig:seeds_cond_conv__nouns_and_verbs} (left) shows $\condexpconvergence{\nounset}$ for the $410m$ parameter model.
All three noun types---regular singular (NN), regular plural (NNS) and proper singular (NNP)---are roughly equally challenging for the model and their final convergence values are either similar or lower than the starting point at training step $0$.

\paragraph{Differences in Verbs.}
\Cref{fig:seeds_cond_conv__nouns_and_verbs} (right) shows the conditional convergence for the tokens where a verb is being predicted. Unlike nouns, verbs generally converge more throughout training. However, gerund or present participle (VBG) and past participle verbs (VBN) converge less than 3rd person singular present (VBZ), past tense (VBD) and base form verbs (VB). The latter might be simpler forms that could be more easily derived from context. 

\newpage

\section{Predicting Convergence with a Linear Regression}
\label{app:linear_regression}

To explore which factors influence convergence the most, we follow \citep{chang-etal-2024-characterizing} in performing a linear regression analysis. 
Our model consists of fitting the equation:
\begin{align}
    \label{eq:linear-regression}
    \convergencefunc(\context) \approx & \ \alpha \cdot \log(\text{freq}(\subword_t)) \\ 
    &\quad + \sum_{p \in \text{ P.o.S. tags}} \beta_{p} \cdot \mathbf{1}_{\text{tag}(\subword_t)=p}  \nonumber \\ 
    &\quad + \sum_{p \in \text{ P.o.S. tags}} \gamma_{p} \cdot \mathbf{1}_{\text{tag}(\subword_{t-1})=p}  \nonumber \\ 
    &\quad + \sum_{m \in \text{Model Sizes}} \delta_m \cdot \mathbf{1}_{m'=m} \nonumber 
\end{align}
across contexts $\context \in \dataset$. 
We fit one such model per training step for all analysed model sizes $m'$.
This allows us to analyse the influence of our different conditional parameters across the training process.

\begin{figure}[h]
    \begin{subfigure}[b]{0.5\linewidth}
        \centering
        \includegraphics[trim={1cm 0 2cm 0},clip,width=\linewidth]{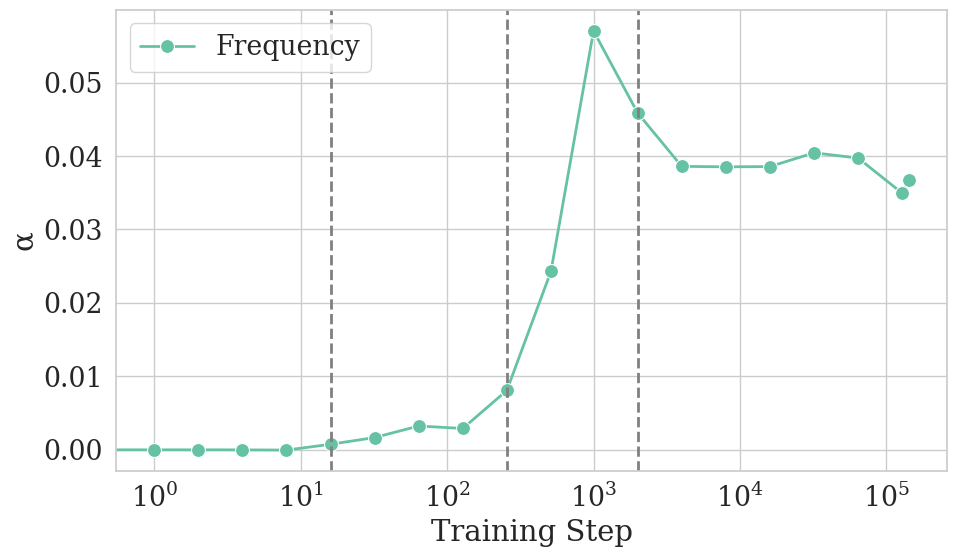}
    \end{subfigure}%
    \begin{subfigure}[b]{0.5\linewidth}
        \centering
        \includegraphics[trim={1cm 0 2cm 0},clip,width=\linewidth]{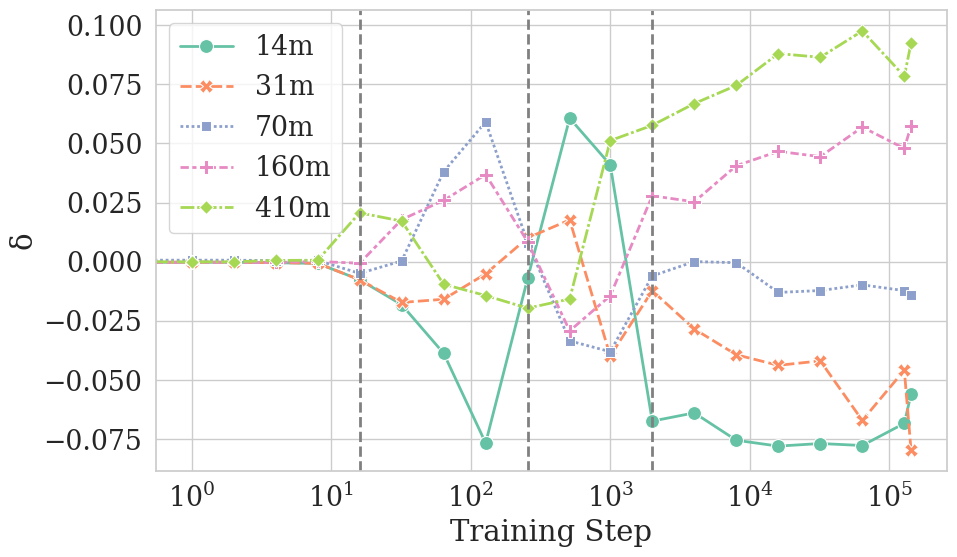}
    \end{subfigure}%
    \caption{Linear regression's coefficients for frequency (left) and model size (right) when predicting convergence: $\convergencefunc(\context)$.}
    \label{fig:linear_regression}
\end{figure}

Frequency (measured by the fitted parameter $\alpha$) emerges as a significant factor influencing KL divergence, exhibiting a distinct pattern over the course of training (see \cref{fig:linear_regression}, left). 
Initially, its influence is negligible, remaining close to zero. Around step $256$, frequency begins to play a stronger role, positively influencing $\convergencefunc$ ($\alpha \approx 0.05$), indicating that frequent tokens stabilise earlier and exhibit lower KL divergence across random seeds. 
Beyond this point, the correlation slightly increases again, suggesting that while frequent tokens converge more quickly, their stabilisation process becomes less distinct as training progresses. 
This trajectory aligns with our observations of token convergence by frequency in \cref{fig:seeds_cond_conv}.

The influence of model size on $\convergencefunc$ initially shows an unstable pattern, but later a trend emerges where larger models increase $\convergencefunc$, while smaller models decrease $\convergencefunc$ relatively (see \cref{fig:linear_regression}, right). 
It should be noted that while the learning rate is the same for the 3 smallest models, their influence on convergence is not.
This suggests that the learning rate warm-up (see \cref{fig:learning_rates}) is not the only determinant of these model convergence patterns.

\section{Conditional Cross-entropies} \label{app:cond_cross_entropies}

\begin{figure*}
    \centering
    \begin{subfigure}[b]{0.33\linewidth}
        \centering
        \includegraphics[trim={0 0 0 0},clip,width=\linewidth]{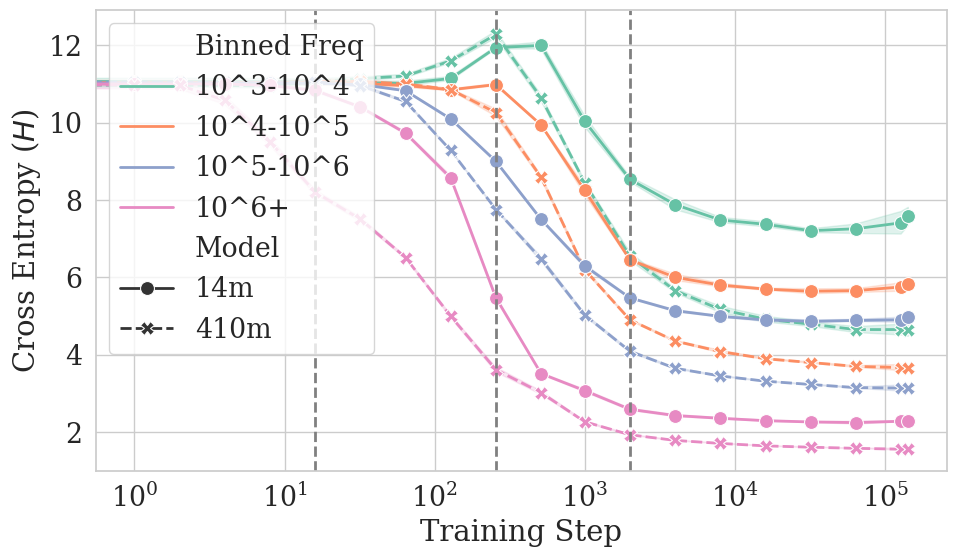}
    \end{subfigure}%
    \hspace{40pt}%
    \begin{subfigure}[b]{0.33\linewidth}
        \centering
        \includegraphics[trim={0 0 0 0},clip,width=\linewidth]{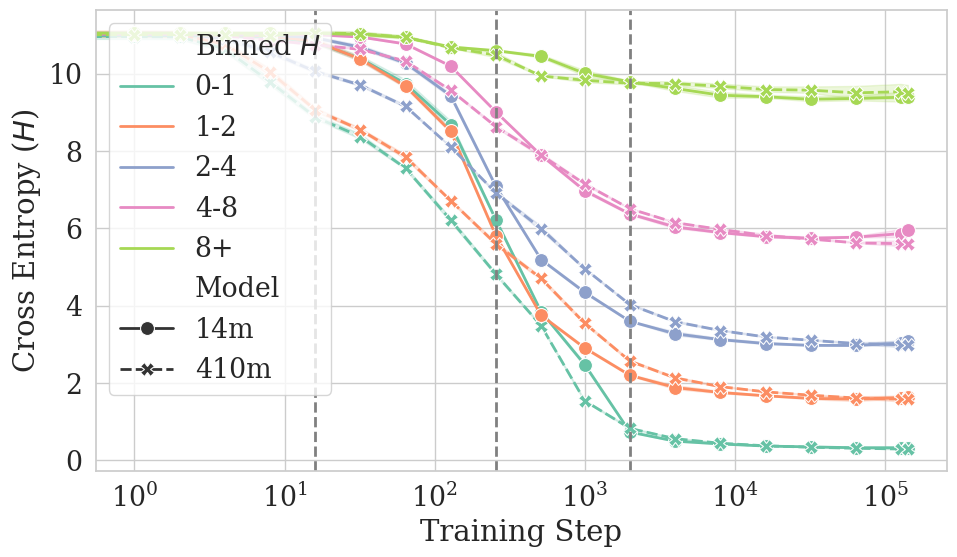}
    \end{subfigure}
    \begin{subfigure}[b]{0.33\linewidth}
        \centering
        \includegraphics[trim={0 0 0 0},clip,width=\linewidth]{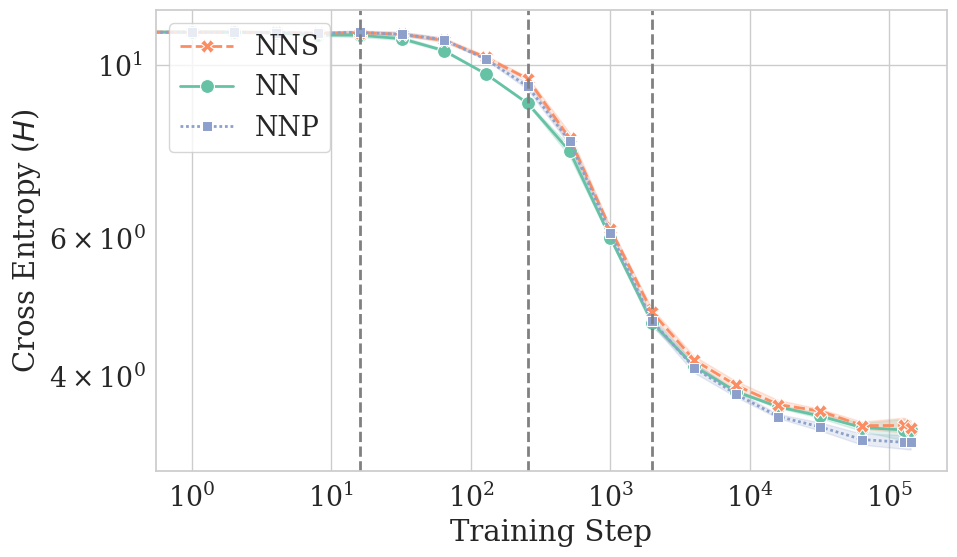}
    \end{subfigure}%
    \begin{subfigure}[b]{0.33\linewidth}
        \centering
        \includegraphics[trim={0 0 0 0},clip,width=\linewidth]{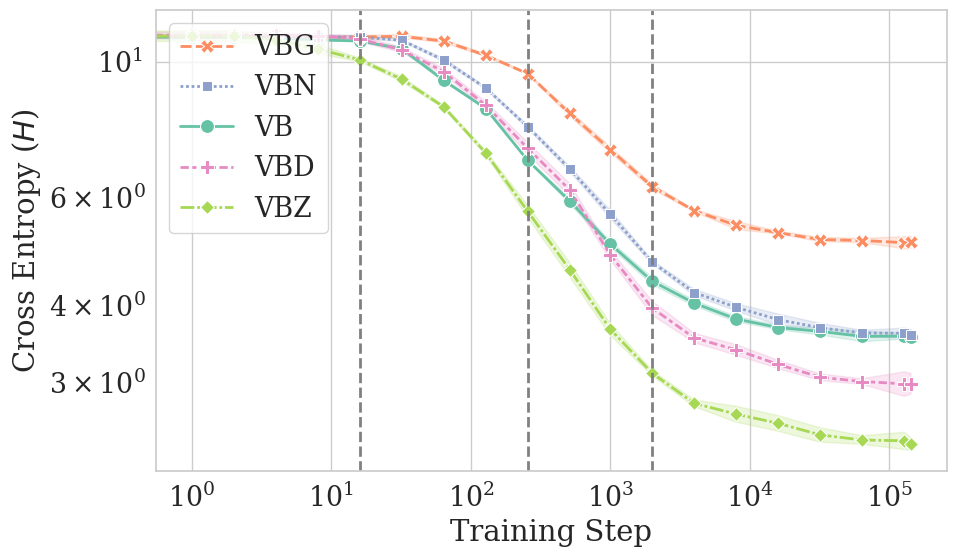}
    \end{subfigure}%
    \begin{subfigure}[b]{0.33\linewidth}
        \centering
        \includegraphics[trim={0 0 0 0},clip,width=\linewidth]{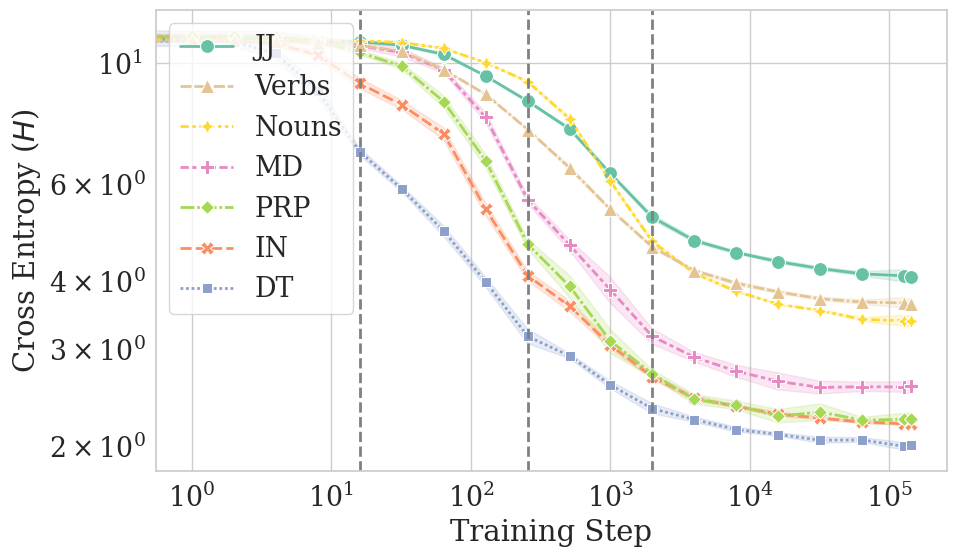}
    \end{subfigure}%
    \caption{
    Conditional cross-entropy ($H$) of models with $1\sigma$ confidence intervals. Conditioning property: 
    (top-left) Frequency; (top-right) Surprisal at end of training;
    (bottom-left) Nouns' PoS; (bottom-center) Verbs' PoS; (bottom-right) Other PoS.}
    \label{fig:cond_cross_entropies}
\end{figure*}

\Cref{fig:cond_cross_entropies} presents conditional cross-entropies for our analysed models and conditioning properties.

\section{Tokens ranked by KL}

\Cref{tab:kl_text_samples} provides a list of tokens with high, low and medium KL respectively. We sampled the tokens exclusively from the natural language text and ignored the code snippet in our sample.

\newcommand{\mytoken}[1]{\texttt{#1}}

\begin{table*}
    \centering
    \small
    \begin{tabular}{llllll}
        \toprule
        \textbf{Token (Low)} & \textbf{KL (Low)} & \textbf{Token (Mid)} & \textbf{KL (Mid)} & \textbf{Token (High)} & \textbf{KL (High)} \\
        \midrule
        \mytoken{ptions} & -0.012750 & \mytoken{)} & -0.411134 & \mytoken{OH} & -1.317826 \\
        \mytoken{ied} & -0.038621 & \mytoken{\_conv} & -0.411211 & \mytoken{ley} & -1.332273 \\
        \mytoken{oration} & -0.052642 & \mytoken{\_known} & -0.411488 & \mytoken{agn} & -1.472763 \\
        \mytoken{isexual} & -0.058059 & \mytoken{\_visual} & -0.411673 & \mytoken{hor} & -1.537567 \\
        \mytoken{ll} & -0.060047 & \mytoken{ify} & -0.411950 & \mytoken{Ed} & -1.572812 \\
        \mytoken{'t} & -0.060089 & \mytoken{en} & -0.412472 & \mytoken{\_=} & -1.690981 \\
        \mytoken{N} & -0.060991 & \mytoken{]} & -0.412764 & \mytoken{var} & -1.712604 \\
        \mytoken{\_About} & -0.065953 & \mytoken{\_founded} & -0.413144 & \mytoken{OR} & -1.760247 \\
        \mytoken{\_am} & -0.066206 & \mytoken{virtual} & -0.414356 & \mytoken{äll} & -2.286940 \\
        \mytoken{ters} & -0.067074 & \mytoken{\_whether} & -0.414530 & \mytoken{LEY} & -2.874433 \\
        \bottomrule
    \end{tabular}
    \caption{Representative tokens with the lowest, highest and medium KL divergence. `\_' represents a whitespace.}
    \label{tab:kl_text_samples}
\end{table*}

\section{Results Conditioning on Context Token}
\begin{figure*}
    \centering
    \begin{subfigure}[b]{0.33\linewidth}
        \centering
        \includegraphics[trim={0 0 0 0},clip,width=\linewidth]{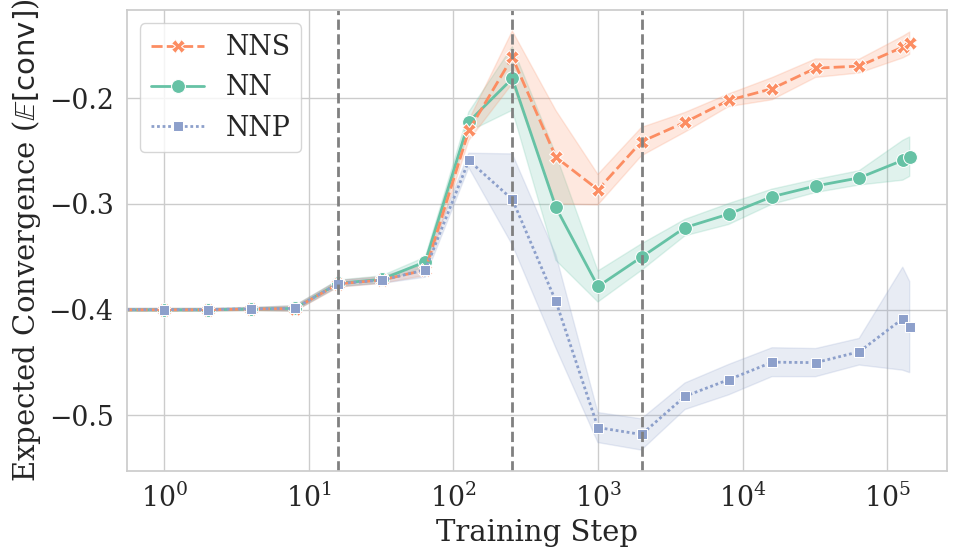}
    \end{subfigure}%
    \begin{subfigure}[b]{0.33\linewidth}
        \centering
        \includegraphics[trim={0 0 0 0},clip,width=\linewidth]{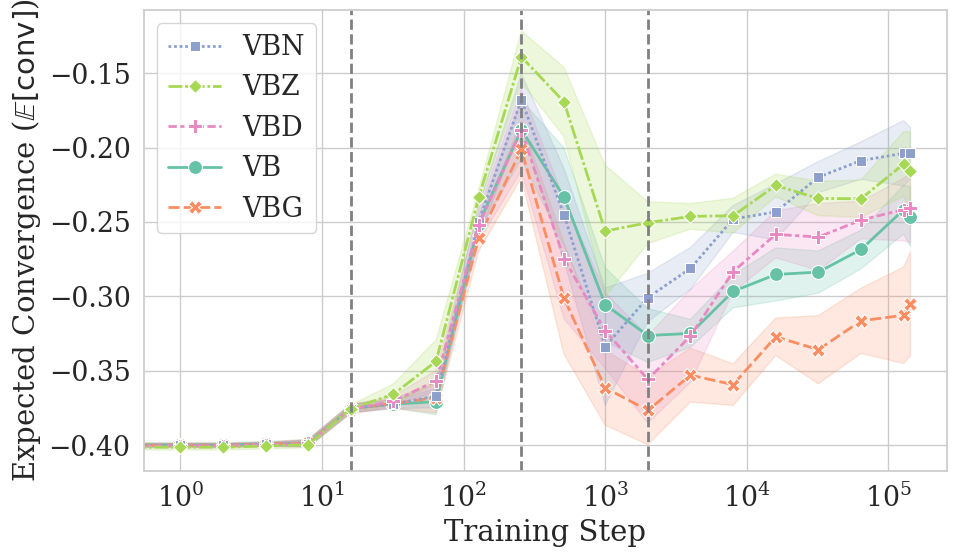}
    \end{subfigure}%
    \begin{subfigure}[b]{0.33\linewidth}
        \centering
        \includegraphics[trim={0 0 0 0},clip,width=\linewidth]{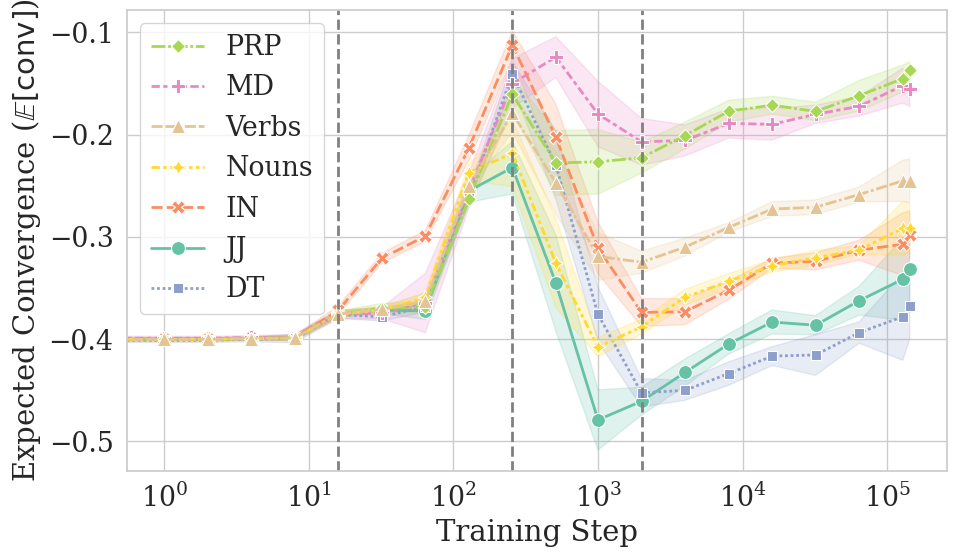}
    \end{subfigure}%
    \caption{
    $\condexpconvergence{\vocab_{t-1}}$ of $410m$ models with $1\sigma$ confidence intervals. Conditioning property on context token ($\subword_{t-1}$):
    (left) Nouns' PoS; (center) Verbs' PoS; (right) Other PoS.}
    \label{fig:cond_kl_context}
\end{figure*}

\begin{figure*}
    \centering
    \begin{subfigure}[b]{0.33\linewidth}
        \centering
        \includegraphics[trim={0 0 0 0},clip,width=\linewidth]{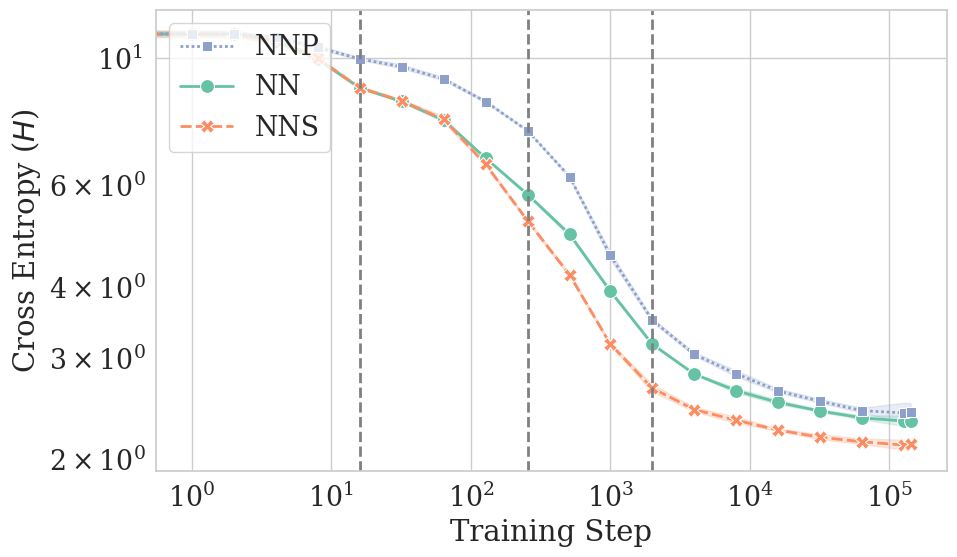}
    \end{subfigure}%
    \begin{subfigure}[b]{0.33\linewidth}
        \centering
        \includegraphics[trim={0 0 0 0},clip,width=\linewidth]{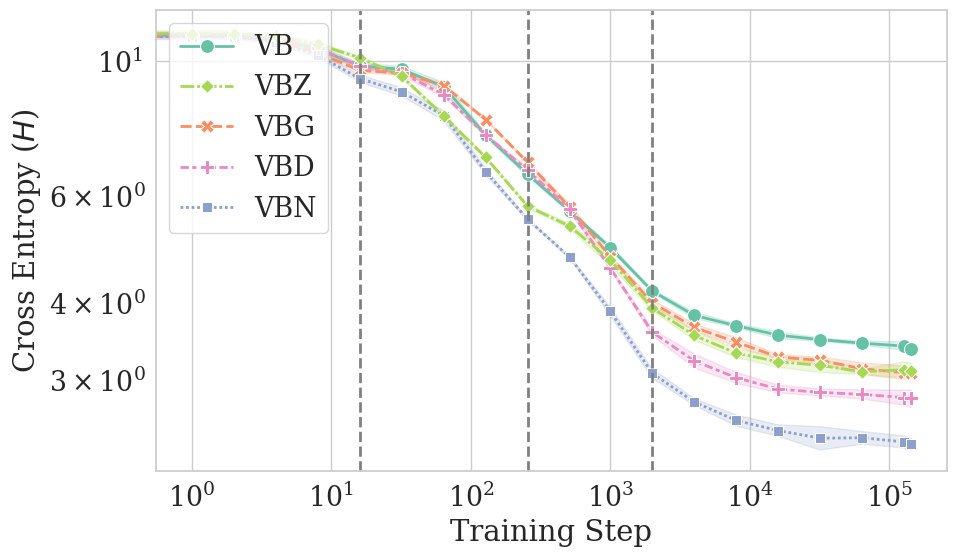}
    \end{subfigure}%
    \begin{subfigure}[b]{0.33\linewidth}
        \centering
        \includegraphics[trim={0 0 0 0},clip,width=\linewidth]{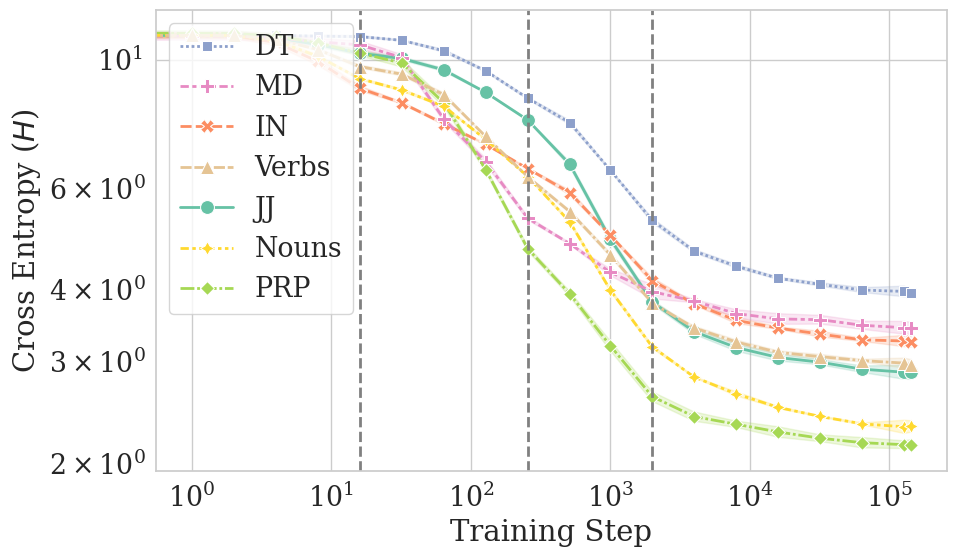}
    \end{subfigure}%
    \caption{
    Conditional cross-entropy ($H$) of $410m$ models with $1\sigma$ confidence intervals. Conditioning property on the final context token ($\subword_{t-1}$):
    (left) Nouns' PoS; (center) Verbs' PoS; (right) Other PoS.}
    \label{fig:cond_cross_entropies_context}
\end{figure*}

\Cref{fig:cond_kl_context} shows the $\condexpconvergence{\vocab_{t-1}}$ of the $410m$ models conditioned on properties of the last token in the context ($\subword_{t-1}$); similarly to \cref{fig:seeds_cond_conv} for the predicted token. 
Similarly, \cref{fig:cond_cross_entropies_context} shows the conditional cross-entropy, conditioned on properties of the last token in context ($\subword_{t-1}$), akin to \cref{fig:cond_cross_entropies}.

\end{document}